\documentclass[letterpaper, 10 pt, conference]{ieeeconf}  

\IEEEoverridecommandlockouts                              

\usepackage[T1]{fontenc}
\usepackage[utf8]{inputenc}

\usepackage{graphics} 
\usepackage{graphicx}
\usepackage{tabularx}
\newcolumntype{C}{>{\centering\arraybackslash}X}
\usepackage[font=footnotesize,labelfont=bf,labelsep=colon]{caption}
\usepackage{epsfig} 
\usepackage{amsmath} 
\usepackage{amssymb}  
\usepackage{multirow}
\usepackage{booktabs}
\usepackage[table]{xcolor}
\usepackage{url}

\usepackage{threeparttable}
\definecolor{linklightblue}{RGB}{70,145,210}
\usepackage[colorlinks=true,urlcolor=linklightblue,linkcolor=black,citecolor=black]{hyperref}

\title{\LARGE \bf
Feeling the Unexpected: ResTacVLA for Contact-Rich Manipulation via Residual Tactile Representation
}

\author{Pengwei Zhang$^{1,2}$, Bin Xie$^{3}$, Xinpan Meng$^{1}$, Xinyu Guo$^{2}$,\\
Ce Hao$^{2}$, Fang Deng$^{2}$, Long Cheng$^{1,*}$, and Tiancai Wang$^{3}$%
\thanks{This work was supported in part by the Brain Science and Brain-like Intelligence Technology -- National Science and Technology Major Project under Grant 2025ZD0215600, in part by the National Natural Science Foundation of China under Grants U25A20475 and 62333023, in part by the Beijing Municipal Natural Science Foundation under Grants F2024201068 and L243014, in part by the CAS Project for Young Scientists in Basic Research under Grant YSBR-034, in part by the Zhongguancun Academy under Grant 20240307, and in part by the Fundamental Research Funds for the Central Universities.}%
\thanks{$^{1}$Pengwei Zhang, Xinpan Meng, and Long Cheng are with the School of Artificial Intelligence, University of Chinese Academy of Sciences, Beijing 100049, China, and are also with the Institute of Automation, Chinese Academy of Sciences, Beijing 100190, China.}%
\thanks{$^{2}$Pengwei Zhang, Ce Hao, Xinyu Guo, and Fang Deng are with Zhongguancun Academy, Beijing 100094, China.}%
\thanks{$^{3}$Bin Xie and Tiancai Wang are with Dexmal, Beijing 100096, China.}%
\thanks{$^{*}$Long Cheng is the corresponding author ({\tt\small longcheng.ia.ac.cn}).}%
}

\begin{document}

\maketitle
\thispagestyle{empty}
\pagestyle{empty}

\begin{abstract}

        \textbf{Tactile perception is indispensable for contact-rich manipulation, yet integrating it into Vision-Language-Action (VLA) models often induces \textit{modality collapse}, where high-bandwidth visual features overshadow sparse tactile cues.} Inspired by \textbf{Predictive Coding}---a neural mechanism where the brain attenuates predictable inputs to prioritize surprising stimuli---we propose \textbf{ResTacVLA}. Rather than treating tactile data as raw input, we \textbf{reformulate} it as a \textbf{Residual Tactile Representation} capturing the discrepancy between visual priors and physical sensations. By filtering out visually predictable dynamics, this formulation transforms sparse tactile signals into \textbf{dense, high-value information gain}, thereby inherently resolving the bandwidth mismatch. These residuals are discretized through a \textbf{Vector Quantized (VQ)} bottleneck into \textbf{Latent Contact Primitives} that capture critical events missed by vision. \textbf{Analogous to the neural surprise signal, we leverage the uncertainty of the visual prior to adaptively gate tactile integration, prioritizing residuals specifically during visually unreliable phases to explicitly prevent visual dominance.} Experimental results show that ResTacVLA consistently outperforms all baselines on a diverse set of contact-rich manipulation tasks, while remaining robust to unexpected dynamic disturbances. Project website: \url{https://awilekong.github.io/ResTacVLA/}.

\end{abstract}

\section{INTRODUCTION}

Empowered by large-scale embodied pretraining, robotic manipulation policies developed by Vision-Language-Action models (VLAs) have recently emerged as a dominant paradigm. However, current VLAs~\cite{black2024pi0, black2025pi05} are primarily vision-centric, relying predominantly on visual perception. Although many manipulation tasks can be completed nearly perfectly using this approach, a critical bottleneck remains in contact-rich manipulation tasks—such as precision insertion~\cite{Heo2023FurnitureBenchRR}, threaded assembly~\cite{Noseworthy2024FORGEFE}, and surface wiping~\cite{Hou2024AdaptiveCP}. While these tasks are relatively straightforward for humans, the visual-centric approach often struggles with severe occlusions or lacks the fidelity required to resolve fine-grained physical dynamics. The primary limitation lies in the absence of mastery over tactile perception, which is indispensable for grounding physical interactions.

\begin{figure}[t]
  \centering
  \includegraphics[width=\columnwidth]{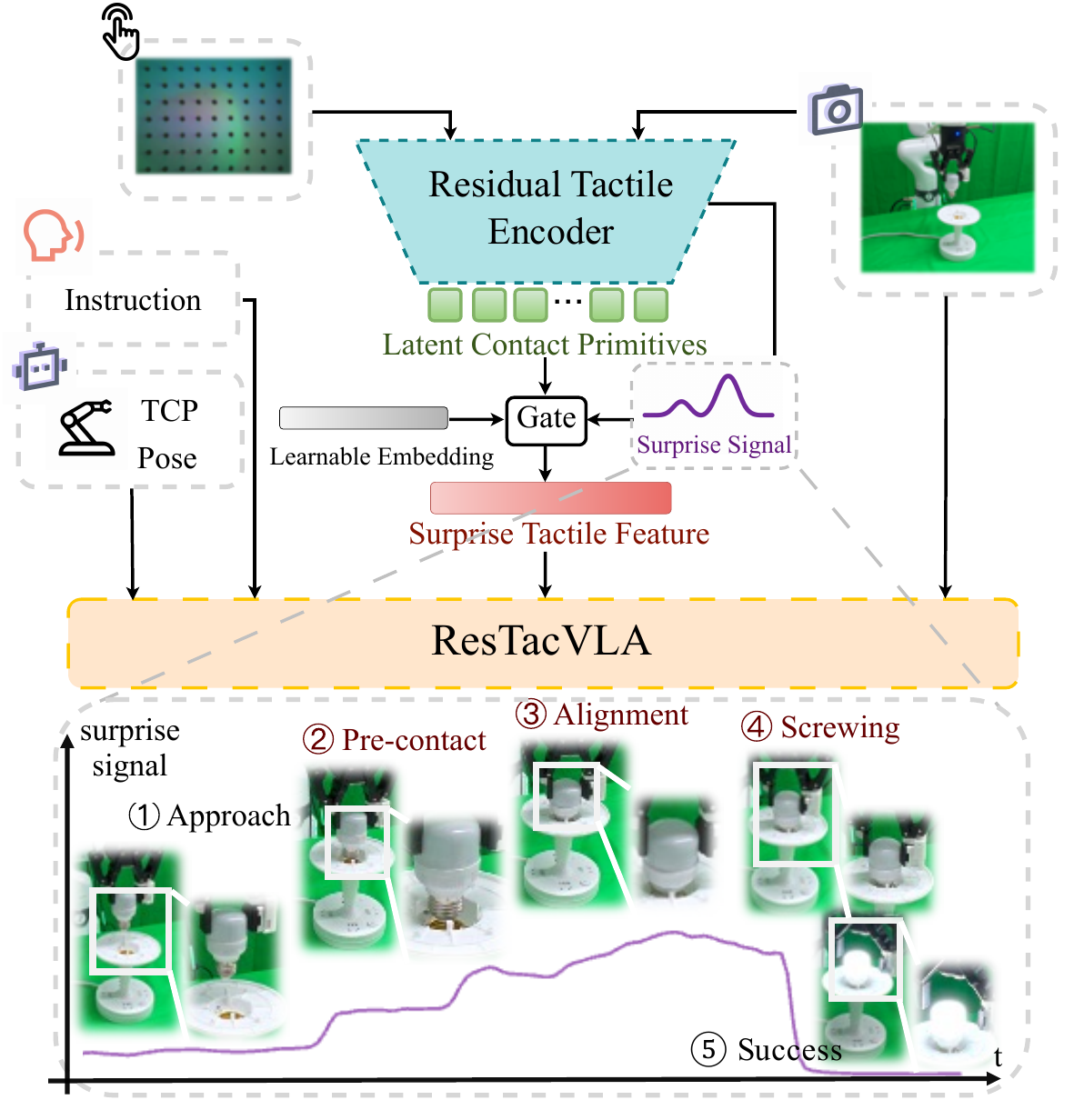}
  \caption{\textbf{Predictive Coding-Inspired Residual Tactile Fusion in ResTacVLA.} The Residual Tactile Encoder extracts high-information-gain representations by modeling the discrepancy between visual priors and physical sensations. A surprise signal drives a Surprise-Aware Gate to adaptively amplify tactile cues during contact-critical phases while suppressing redundant tactile noise in free-space motion.}
  \label{fig:teaser}
\vspace{-5mm}
\end{figure}

A naive approach to incorporating tactile perception is to simply treat it as an additional modality, directly projecting tactile inputs into a shared feature space~\cite{jones2025beyond, bi2025vlatouch, zhang2025vtla}. However, in practice, a phenomenon known as `Modality Collapse'~\cite{Chaudhuri2025ACL} occurs, where the high-bandwidth, continuous visual stream naturally overshadows the event-driven, temporally sparse tactile signals~\cite{Chen2025MultiModalMV, Chen2025ImplicitRDPAE}. By overlaying the missing tactile modality during pretraining, VLAs tend to disregard the `quiet' tactile cues in favor of the `loud' visual features. The inefficiency of this native approach compels us to find a new way to empower VLAs with tactile perception. Recent research in cognitive neuroscience has shown that biological systems address the `Modality Collapse' through Predictive Coding~\cite{Kilteni2022Predictive, Rao1999PredictiveCI, Friston2010TheFP}. Specifically, instead of processing all modality inputs equally, the brain generates top-down predictions of expected sensory states and attenuates predictable inputs, focusing attention on the `Unexpected'—the surprise arising from deviations from expectations. For example, humans cannot tickle themselves because the brain predicts the sensory consequences of self-generated movements and suppresses the expected signal, prioritizing only unexpected external stimuli~\cite{Blakemore1998CentralCO}. This mechanism allows organisms to filter out redundant information and respond quickly to anomalies. Naturally, the question arises: How can we enable current VLAs to feel the unexpected?

To address this challenge, we propose a Residual Tactile Vision-Language-Action method, called ResTacVLA (Fig.~\ref{fig:teaser}), a novel framework that explicitly models the concept of `feeling the unexpected' for contact-rich manipulation tasks. Rather than competing with the high-bandwidth visual modality, ResTacVLA reformulates tactile feedback as a Residual Tactile Representation—a quantitative measure of surprise relative to visual priors. To generate this representation, we introduce the Cross-Modal Predictor (CMP), which distills the discrepancy between visual expectations and physical reality, transforming sparse tactile signals into dense, high-value information. These residuals are then discretized through vector quantization into Latent Contact Primitives that capture intrinsic physical events (e.g., unexpected collisions) that vision fails to perceive. To process the Residual Tactile Representation, we employ a Surprise-Aware Gate (SAG) that adaptively regulates tactile integration, suppressing noise when vision is reliable and explicitly amplifying the tactile pathway during physically ambiguous phases.

In order to evaluate ResTacVLA, we selected five real-world contact-rich tasks, including precision insertion, screwing, and surface wiping. Experimental results demonstrate that ResTacVLA achieves state-of-the-art performance, significantly outperforming both standard VLA baselines and naive tactile fusion strategies, while exhibiting superior robustness against dynamic physical disturbances.

Our contributions are summarized as follows:

\begin{itemize}
        \item We propose ResTacVLA, a biologically inspired framework that integrates tactile feedback through Residual Tactile Representations. By filtering out visual redundancy, we transform tactile signals into dense information gain, effectively mitigating the modality imbalance problem.
        \item We design a Cross-Modal Predictor (CMP) that encodes tactile signals into Latent Contact Primitives with high information gain, and a Surprise-Aware Gate (SAG) is introduced to adaptively modulate tactile injection based on task-phase characteristics. 
        \item Through experiments on five challenging tasks, ResTacVLA achieves up to 86.7\% task success and improves average performance by 34.6\% over baselines, while maintaining strong robustness against dynamic disturbances.
\end{itemize}

\begin{figure*}[t]
\vspace{1mm}
  \centering
  \includegraphics[width=.9\textwidth]{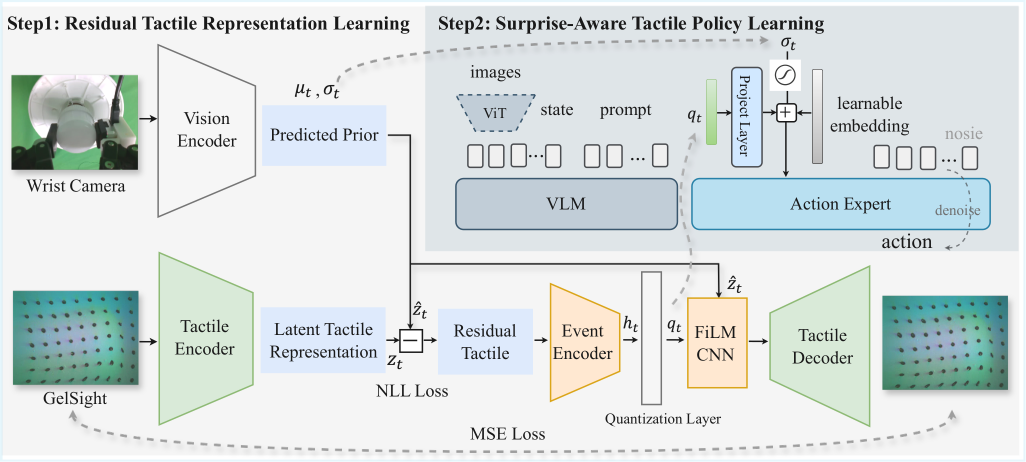}
  \caption{\textbf{Overall architecture of ResTacVLA.} (a) \textbf{Step 1: Residual Tactile Representation Learning.} The Cross-Modal Predictor (CMP) is pre-trained to estimate tactile latents from wrist-camera observations, extracting residual representations that capture the discrepancy between visual priors and physical sensations. These residuals are discretized into Latent Contact Primitives via a VQ bottleneck. (b) \textbf{ Step 2: Surprise-Aware Tactile Policy Learning.} With the CMP frozen, the prediction uncertainty $\sigma_t$ drives a Surprise-Aware Gate (SAG) to adaptively inject contact primitives into the action expert, conditioning a flow matching policy for contact-rich manipulation.}
  \label{fig:method}
\vspace{-6mm}
\end{figure*}

\section{RELATED WORK}

\subsection{Vision-Language-Action Models in Contact-Rich Manipulation}

The emergence of Vision-Language-Action (VLA) learning has revolutionized robotic manipulation by bridging the gap between large-scale visual perception and natural language reasoning~\cite{black2024pi0,Shi2025MemoryVLAPM}. Prominent frameworks, ranging from transformer-based action predictors~\cite{Brohan2022RT1RT,Brohan2023RT2VM} to open-weight foundation models~\cite{black2024pi0,black2025pi05}, harness extensive datasets of aligned visual observations, linguistic instructions, and kinematic trajectories. Leveraging scaling laws in sequence modeling, these systems demonstrate strong generalization across semantically diverse manipulation scenarios. However, their capabilities remain constrained to high-level semantic understanding. While VLAs effectively determine \textit{what} to do, they lack mechanisms to reason about \textit{how} to perform contact-rich interactions---regulating forces, adapting to compliance, or recovering from collisions~\cite{bi2025vlatouch,zhang2025vtla,hao2025tla}. This `tactile-blind' nature renders them vulnerable when occlusion obscures geometry or when physical dynamics (friction, deformation) cannot be inferred from vision alone, necessitating direct tactile feedback for robust execution.

\subsection{Representation Learning of Vision-Based Tactile Sensing}

Constructing robust representations for high-dimensional tactile data is a prerequisite for effective physical interaction. Self-supervised methods---including masked modeling~\cite{higuera2024sparsh} and compact task-agnostic encoding~\cite{xu2024unit}---have established a solid foundation by extracting generalizable physical features from raw tactile streams without extensive annotation. To further bridge the semantic gap between touch and vision, cross-modal contrastive approaches such as VITaL~\cite{zhao2025vital} and Beyond Sight~\cite{jones2025beyond} align tactile embeddings with visual or linguistic latent spaces by maximizing inter-modal mutual information. Although effective at exploiting shared information, this alignment paradigm inadvertently suppresses the critical residual information unique to touch---specifically, the fine-grained contact dynamics that vision cannot predict. For contact-rich manipulation, we argue that task success hinges precisely on this residual. Consequently, rather than prioritizing cross-modal alignment, our approach explicitly models the tactile residual, capturing the discrepancy between visual expectation and physical reality to maximize information gain.

\subsection{Tactile Integration in Generalist Policies}

Strategies for integrating tactile sensing into VLA frameworks typically fall into two categories. One line of work augments policies with global force feedback~\cite{yu2025forcevla,zhang2025tavla}, utilizing wrist-mounted Force/Torque (F/T) sensors or joint-torque estimation. Although effective for collision detection or payload monitoring, these modalities fundamentally aggregate complex interaction dynamics into a single resultant vector, thereby forfeiting the spatial richness necessary to capture local contact geometry and distributed pressure patterns.

In contrast, visuotactile-based VLA policies~\cite{bi2025vlatouch,zhang2025vtla,hao2025tla,huang2025tactilevla} seek to exploit the high-resolution capabilities of vision-based tactile sensors. However, the prevailing paradigm treats tactile frames merely as `auxiliary visual features', fusing them with scene observations through standard Transformer architectures. We argue that this naive integration induces `Modality Collapse': the high-bandwidth, continuous stream of visual data naturally dominates sparse, event-driven tactile signals. Without explicit mechanisms to filter visual redundancy, such models tend to disregard tactile feedback in favor of dominant visual priors, thereby failing to leverage the distinctive information gain provided by touch during critical contact phases.

\section{METHODOLOGY}
\textbf{Problem Formulation.} Similar to other vision-centric VLA policies, the objective of the tactile-aware policy $\pi$ is to improve the performance on contact-rich tasks by mapping inputs—including visual, tactile, and language instructions—into low-level actions. Formally, at timestep $t$, the policy processes both the observation $O_t$ and the language instruction $L$, outputting a low-level action sequence $A_t = \{a_t, a_{t+1}, \ldots, a_{t+H-1}\}$, i.e., $\pi(A_t | O_t, L)$. The visual modality, coming from the robot's cameras, consists of base, side, and wrist visual inputs: $V_t^{base}$, $V_t^{side}$, and $V_t^{wrist}$; the tactile modality is represented by $I_t^{tac}$; and the proprioceptive state is $s_t \in \mathbb{R}^{7}$. These modalities are collectively denoted as $O_t = \{V_t^{base}, V_t^{side}, V_t^{wrist}, I_t^{tac}, s_t\}$. The language instructions $L$ describe the tasks that the policy is required to perform.

\subsection{Overview}
ResTacVLA is an end-to-end multimodal robotic policy designed for contact-rich manipulation, with its overall pipeline illustrated in Fig.~\ref{fig:method}. Its core mechanism comprises two key components: a Cross-Modal Predictor (CMP) and a Surprise-Aware Gate (SAG). Rather than naively fusing raw sensory streams, the CMP reformulates tactile feedback into a Residual Tactile Representation that captures the deviation of physical reality from visual anticipation. Grounded in Predictive Coding, the CMP distills visual redundancy into discrete Latent Contact Primitives that encode intrinsic contact dynamics. These primitives are then adaptively modulated by the SAG and injected into the action denoising process, allowing the policy to explicitly refine action trajectories based on unexpected physical dynamics, while suppressing tactile noise during visually reliable phases. To acquire robust contact representations prior to policy learning, the CMP is first pre-trained independently on multi-task interaction data, and is subsequently frozen while the full tactile-augmented policy is fine-tuned end-to-end.

Building upon the $\pi_{0.5}$ framework~\cite{black2025pi05}, the tactile-augmented policy integrates vision, language, proprioception, and high-resolution tactile feedback to generate actions through a conditional flow-matching model~\cite{lipman2022flow}. Visual inputs from multiple RGB cameras and task instructions, along with proprioceptive states, are encoded by a pretrained vision-language model, e.g., PaliGemma~\cite{beyer2024paligemma} into contextual embeddings. These embeddings, combined with the modulated contact primitives, condition an iterative denoising process that predicts the action trajectory.

\subsection{Residual Tactile Representation Learning}
\label{sec:residual_tactile}

\subsubsection{Cross-Modal Predictor and Residual Tactile Extraction}
We introduce the Cross-Modal Predictor (CMP) to estimate the expected latent tactile representation exclusively from wrist camera observations. Specifically, a Vision Encoder processes $V_{t}^{wrist}$ via a learnable ResNet-18 backbone followed by an MLP head, yielding a predicted mean $\mu_{t}\in\mathbb{R}^{3\times H^{\prime}\times W^{\prime}}$ and a scalar standard deviation $\sigma_{t}\in\mathbb{R}$. These parameters characterize a Gaussian distribution over the predicted tactile latent $\hat{z}_{t}$. Concurrently, a tactile encoder, based on UniT~\cite{xu2024unit}, projects the actual tactile image $I_{t}^{tac}$ into the same latent space, producing $z_{t}\in\mathbb{R}^{3\times H^{\prime}\times W^{\prime}}$. The residual tactile, defined as $r_{t}=z_{t}-\hat{z}_{t}$, isolates physical sensory components unanticipated by the visual modality. The predictor is optimized using a weighted Negative Log-Likelihood (NLL)~\cite{kendall2017uncertainties} objective, which promotes accurate prediction while explicitly modeling aleatoric uncertainty:
\begin{equation}
    \mathcal{L}_{pred} = \lambda_{\sigma} \log \sigma_{t}^{2} + \frac{\|z_{t} - \mu_{t}\|^{2}}{\sigma_{t}^{2}},
    \label{eq:pred_loss}
\end{equation}
where hyperparameter $\lambda_{\sigma}$ regulates the variance penalty to prevent uncertainty collapse. Crucially, $\sigma_{t}$ functions as a principled measure of cross-modal surprise, which is subsequently utilized by the SAG (Sec.~\ref{sec:policy_learning}) to adaptively modulate the contribution of tactile information.

\subsubsection{Latent Contact Primitives via VQ}
An event encoder $f_{\phi}$ processes the residual tactile $r_{t}$ and aggregates features into a global event vector $h_{t}=f_{\phi}(r_{t})\in\mathbb{R}^{D}$ employing convolutional residual blocks followed by global max pooling. The $h_{t}$ is subsequently discretized into a Latent Contact Primitive $q_{t}$ via a Vector Quantization (VQ)~\cite{oord2017neural} bottleneck containing a learnable codebook $\mathcal{C}=\{c_{k}\}_{k=1}^{K}$. To mitigate codebook collapse, we implement multiple strategies during VQ training: reducing the codebook dimension~\cite{yu2021vitvqgan}, replacing Euclidean distance with cosine similarity~\cite{yu2021vitvqgan}, smoothing codebook updates via exponential moving average (EMA)~\cite{vqvae2}, and periodically reinitializing inactive codebook entries~\cite{zeghidour2021soundstream}. The latent tactile representation is then reconstructed by modulating the predicted prior $\hat{z}_{t}$ with the quantized primitive $q_{t}$ via FiLM~\cite{perez2018film} conditioning. The resulting $\tilde{z}_{t}$ is decoded back into the tactile image space, yielding the reconstructed tactile image $\hat{I}_t$. The complete CMP pipeline is jointly trained on multi-task interaction data by minimizing the following objective:
\begin{equation}
    \mathcal{L}_{CMP} = \mathcal{L}_{rec} + \lambda_{p} \mathcal{L}_{pred} + \mathcal{L}_{vq},
    \label{eq:restac_loss}
\end{equation}
where $\mathcal{L}_{rec} = \| \hat{I}_t - I_t^{tac} \|_2^2$ denotes the mean squared error (MSE) for tactile reconstruction. $\mathcal{L}_{pred}$ is the cross-modal prediction loss (Eq.~\ref{eq:pred_loss}) weighted by hyperparameter $\lambda_p$, and $\mathcal{L}_{vq}$ is the standard VQ commitment loss \cite{yu2021vitvqgan}.
\subsection{Surprise-Aware Tactile Policy Learning}
\label{sec:policy_learning}

The prediction uncertainty $\sigma_{t}$, derived from the CMP, serves as a principled indicator for adaptive modality fusion. An elevated value of $\sigma_{t}$ signifies that the visual system cannot reliably anticipate the current tactile state, implying low cross-modal mutual information and, consequently, high tactile information gain. Conversely, a low $\sigma_{t}$ suggests that tactile feedback is largely redundant relative to visual observations. To leverage this characteristic, we compute a surprise-aware gate $g_{t} \in (0,1)$ as follows:
\begin{equation}
    g_{t} = \text{Sigmoid}(\text{MLP}(\sigma_{t})).
\end{equation}
This gate is then applied to modulate the tactile pathway. Specifically, the Latent Contact Primitive $q_{t}$, obtained from the frozen CMP (Sec.~\ref{sec:residual_tactile}), is projected into the token dimension of the action expert via a linear layer, yielding $p_{t} \in \mathbb{R}^{d}$. Simultaneously, a learnable embedding $e_{0} \in \mathbb{R}^{d}$ is employed as a default `no-contact' token. The final tactile token $e_{t}$ is synthesized via gated interpolation:
\begin{equation}
    e_{t} = g_{t} \cdot p_{t} + (1 - g_{t}) \cdot e_{0}.
\end{equation}
When visual predictions exhibit high confidence (i.e., $g_{t} \rightarrow 0$), the output converges to $e_{0}$, effectively suppressing the tactile pathway. In contrast, under conditions of high prediction uncertainty (i.e., $g_{t} \rightarrow 1$), the contact primitive $p_{t}$ dominates, thereby explicitly injecting physical information into the policy. To integrate tactile cues into action generation without compromising the pre-trained representations of the VLM, we directly concatenate $e_t$ with the noise tokens as input to the action expert. With the pre-trained CMP frozen, the entire tactile-augmented policy is fine-tuned end-to-end using the standard conditional flow-matching objective~\cite{lipman2022flow}.
\begin{figure*}[t]
 \vspace{2mm}
  \centering
  \includegraphics[width=.95\textwidth]{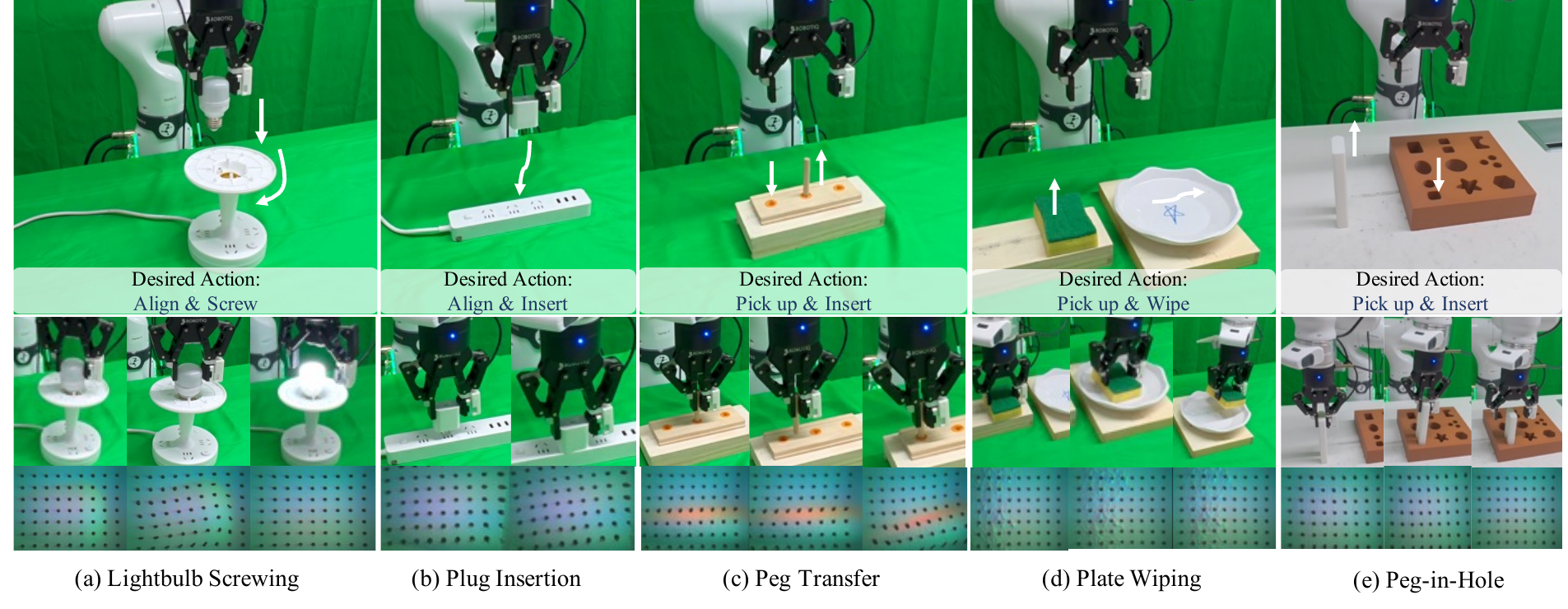}
  \caption{\textbf{Overview of the five contact-rich manipulation tasks designed for evaluation.} (a) \textbf{Lightbulb Screwing}: detecting thread engagement and rotational resistance---dynamics imperceptible to vision alone; (b) \textbf{Plug Insertion}: resolving sub-millimeter positional uncertainty under severe visual occlusion via tactile-guided alignment; (c) \textbf{Peg Transfer}: completing sequential fine-grained manipulation requiring high tactile sensitivity; (d) \textbf{Plate Wiping}: maintaining sustained surface contact through active force compliance to prevent end-effector hovering; (e) \textbf{Peg-in-Hole}: disambiguating contact state under occlusion with tactile precision.}
  \label{fig:tasks}
\vspace{-6mm}
\end{figure*}

\section{EXPERIMENTS}

This section presents a comprehensive suite of real-world experiments to empirically validate ResTacVLA. The evaluation is structured around four core research questions:
\begin{enumerate}
    \item In contact-rich manipulation tasks characterized by sparse tactile modalities, how does the overall effectiveness of ResTacVLA compare to standard baselines and naive tactile fusion strategies?
    \item Do the proposed Residual Tactile Representation and Surprise-Aware Gating play a critical role in enhancing the capabilities of the policy?
    \item Can the Latent Contact Primitives capture meaningful contact events, and the Surprise-Aware Gating adaptively regulate modality importance across different task phases?
    \item Does the policy maintain robustness under unexpected physical perturbations and sensory corruptions?
\end{enumerate}

\subsection{Experimental Setup}

\textbf{Platform.} The experimental platform comprises a Franka Research 3 robotic arm equipped with a Robotiq 2F-85 parallel gripper. A GelSight Mini tactile sensor is mounted on one fingertip of the gripper to capture high-resolution contact geometry. The robot operates within a workspace of dimensions $45 \times 60 \times 40$\,cm. Three Intel RealSense D435 cameras are deployed for visual perception: one positioned frontally, one laterally, and one wrist-mounted near the gripper. All devices are connected to a workstation equipped with an Intel Core i9-10900K CPU and an NVIDIA RTX 4090 GPU, which supports both data collection and policy evaluation.

\textbf{Tasks and Data Collection.} To evaluate ResTacVLA, we curated five tasks that span a spectrum of contact-rich manipulation primitives, as shown in Fig.~\ref{fig:tasks}. These tasks represent distinct physical challenges: \textit{Lightbulb Screwing} necessitates the detection of thread engagement and rotational resistance---dynamics imperceptible to vision alone. \textit{Plug Insertion}, \textit{Peg-in-Hole}, and \textit{Peg Transfer} all impose rigorous sub-millimeter tolerances under severe visual occlusion, mandating tactile-guided alignment to resolve position uncertainty. \textit{Plate Wiping} necessitates active force feedback to guarantee sustained physical contact and effective cleaning, thereby preventing the end-effector from hovering above the surface due to inherent visual depth inaccuracies.

For each task, approximately 100 expert demonstrations are collected, during which the operator receives real-time tactile deformation imagery as tactile feedback. Evaluation is conducted over 25 trials for \textit{Lightbulb Screwing} and \textit{Plug Insertion}, and over 15 trials for \textit{Peg-in-Hole}, \textit{Peg Transfer}, and \textit{Plate Wiping} due to their longer rollout durations, with success determined by task-specific criteria (e.g., electrical continuity for lightbulb screwing, full insertion depth for peg and plug tasks, and effective coverage for wiping). Collectively, these tasks are designed to rigorously probe the capacity of ResTacVLA to handle complex contact interactions, uncertain dynamics, and fine-grained sensorimotor coordination through the integration of vision and tactile modalities.

\textbf{Evaluation Metrics and Baselines.} Model performance is primarily evaluated using the task success rate across all five contact-rich tasks. In addition, for \textit{Lightbulb Screwing}, \textit{Peg-in-Hole}, and \textit{Plug Insertion}, we further decompose evaluation into two phases: (1) \textit{Alignment}---whether the object is correctly positioned at the target---and (2) \textit{Successful Interaction}---whether the task is physically completed (i.e., electrical continuity for the lightbulb, and full insertion depth for the peg and plug). This two-phase metric provides finer-grained insight into potential failure modes.

To comprehensively evaluate ResTacVLA and the proposed CMP and SAG components, we compare it against several carefully selected baselines derived from two foundational architectures: the state-of-the-art $\pi_{0.5}$~\cite{black2025pi05} VLA model and Diffusion Policy~\cite{Chi2023DiffusionPV}. The specific variants include: (1) DP w/o T, a standard Diffusion Policy without tactile input; (2) DP w/ T-ResTac, Diffusion Policy augmented with the proposed Residual Tactile Representation; (3) $\pi_{0.5}$ w/o T, the standard $\pi_{0.5}$ without tactile input; (4) $\pi_{0.5}$ w/ T-ResNet, $\pi_{0.5}$ with tactile images directly encoded by a ResNet-18 backbone; and (5) $\pi_{0.5}$ w/ T-UniT, $\pi_{0.5}$ with tactile images encoded by the pretrained UniT~\cite{xu2024unit} model---a strong VQVAE-based tactile representation with demonstrated effectiveness in contact-rich manipulation. The inclusion of $\pi_{0.5}$ enables comparison with a strong VLA foundation, while the tactile-augmented variants on both $\pi_{0.5}$ and Diffusion Policy are crucial for demonstrating the efficacy of the proposed Residual Tactile Representation over simpler tactile integration approaches such as direct encoding or pretrained feature extraction.

\begin{table*}[t]
\vspace{2mm}
\centering
\renewcommand{\arraystretch}{1.2}
\caption{\textbf{Quantitative Comparison on the Contact-Rich Manipulation Benchmark.}}
\label{tab:main_results}

\begingroup
\renewcommand\TPTminimum{\textwidth}
\begin{threeparttable}

\makebox[\textwidth][c]{%
\resizebox{0.99\textwidth}{!}{%
\begin{tabular}{l!{\vrule width 0.1pt}cccccccc!{\vrule width 0.1pt}c}
\toprule
\rowcolor{gray!8}
\multicolumn{1}{l!{\vrule width 0.4pt}}{\cellcolor{gray!8}\textbf{Method}}
& \textbf{Lightbulb-A}
& \textbf{Lightbulb-I}
& \textbf{Plug-A}
& \textbf{Plug-I}
& \textbf{Peg-A}
& \textbf{Peg-I}
& \textbf{Transfer}
& \multicolumn{1}{c!{\vrule width 0.4pt}}{\cellcolor{gray!8}\textbf{Wiping}}
& \textbf{\textit{Average}} \\
\hline
$\pi_{0.5}$ (Vision Only)  & 28.0  & 8.0  & 36.0 & 20.0 & 46.7 & 40.0 & 26.7 & 20.0 & 28.2 \\
DP w/o T                   & 20.0  & 0.0  & 28.0  & 16.0  & 40.0 & 26.7 & 13.3  & 6.7 & 18.8 \\
$\pi_{0.5}$ w/ T-ResNet    & 16.0  & 0.0  & 32.0 & 24.0 & 40.0 & 40.0 & 20.0 & 13.3 & 23.2 \\
$\pi_{0.5}$ w/ T-UniT      & 28.0 & \underline{12.0} & \underline{40.0} & \underline{32.0} & \underline{73.3} & 53.3 & \textbf{66.7} & \underline{33.3} & \underline{42.3} \\
DP w/ T-ResTac             & \underline{32.0} & \underline{12.0} & 32.0 & 28.0 & 66.7 & \underline{60.0} & 40.0 & \underline{33.3} & 38.0 \\
\hline
\rowcolor{green!3}
\multicolumn{1}{l!{\vrule width 0.4pt}}{\cellcolor{green!3}\textbf{ResTacVLA (Ours)}}
& \textbf{56.0}
& \textbf{32.0}
& \textbf{68.0}
& \textbf{60.0}
& \textbf{86.7}
& \textbf{80.0}
& \underline{60.0}
& \multicolumn{1}{c!{\vrule width 0.4pt}}{\cellcolor{green!3}\textbf{60.0}}
& \textbf{62.8} \\
\bottomrule
\end{tabular}%
}%
}

\begin{tablenotes}[flushleft]
\item \small We report the success rates (\%) and average performance of ResTacVLA against various vision-only and naive tactile fusion baselines across five challenging tasks. Results distinguish between initial alignment (A) and successful physical interaction (I) to highlight the resolution of physical ambiguities. (Bold: best results; Underlined: second-best)
\end{tablenotes}

\end{threeparttable}
\endgroup
\vspace{-4mm}
\end{table*}

\subsection{Main Results}

\textbf{Overall Performance.} As presented in Table~\ref{tab:main_results}, ResTacVLA demonstrates superior performance across all five contact-rich tasks, achieving an average success rate of 62.8\%. This represents a substantial improvement over vision-only baselines, surpassing both the $\pi_{0.5}$ (28.2\%) and the Diffusion Policy (18.8\%) by margins of +34.6\% and +44.0\%, respectively. A phase-wise analysis reveals that while visual policies perform adequately during the \textit{Alignment} phase, their success rates decline sharply during the \textit{Interaction} phase (e.g., in \textit{Plug Insertion}, success drops from 36.0\% to 20.0\%). This performance gap highlights the limitations of visual perception in resolving state ambiguities under occlusion and contact constraints. In contrast, ResTacVLA maintains consistent performance throughout the interaction sequence (68.0\% $\rightarrow$ 60.0\%), effectively leveraging residual tactile feedback to bridge the `Physical Gap' where vision-based policies consistently falter.

\textbf{Efficacy of Residual Tactile Representation.} We further evaluate the contribution of the proposed Residual Tactile Representation by comparing it against alternative tactile integration strategies. Integrating tactile features from a standard ResNet encoder ($\pi_{0.5}$ w/ T-ResNet) yields only marginal improvements over the vision-only baseline, and even degrades performance in complex contact tasks such as \textit{Lightbulb Screwing} (28.0\% → 16.0\%). While integrating a pre-trained SSL model ($\pi_{0.5}$ w/ T-UniT) provides more consistent improvements, the gains remain limited compared to ResTacVLA. This suggests that without effective regulation, high-dimensional tactile signals can act as distractors, introducing sensory interference that disrupts policy decision-making.

Conversely, incorporating the Residual Tactile Representation into the $\pi_{0.5}$ backbone elevates the success rate to 62.8\%. This significant gain confirms that explicitly modeling the residual tactile---representing the deviation between expected and actual sensations---is essential for extracting orthogonal information gain that vision cannot provide. Furthermore, the application of the Residual Tactile Representation to the Diffusion Policy architecture results in a similar +19.2\% improvement, demonstrating the architecture-agnostic effectiveness of our residual representation in tactile integration.


\subsection{Interpretability Analysis}
\label{sec:interpretability}

To validate the hypothesis that the CMP learns meaningful contact semantics and the SAG provides adaptive gating mechanisms, we conduct a qualitative analysis of the internal representations and gating dynamics.

\begin{figure}[h]
  \centering
  \includegraphics[width=.9\columnwidth]{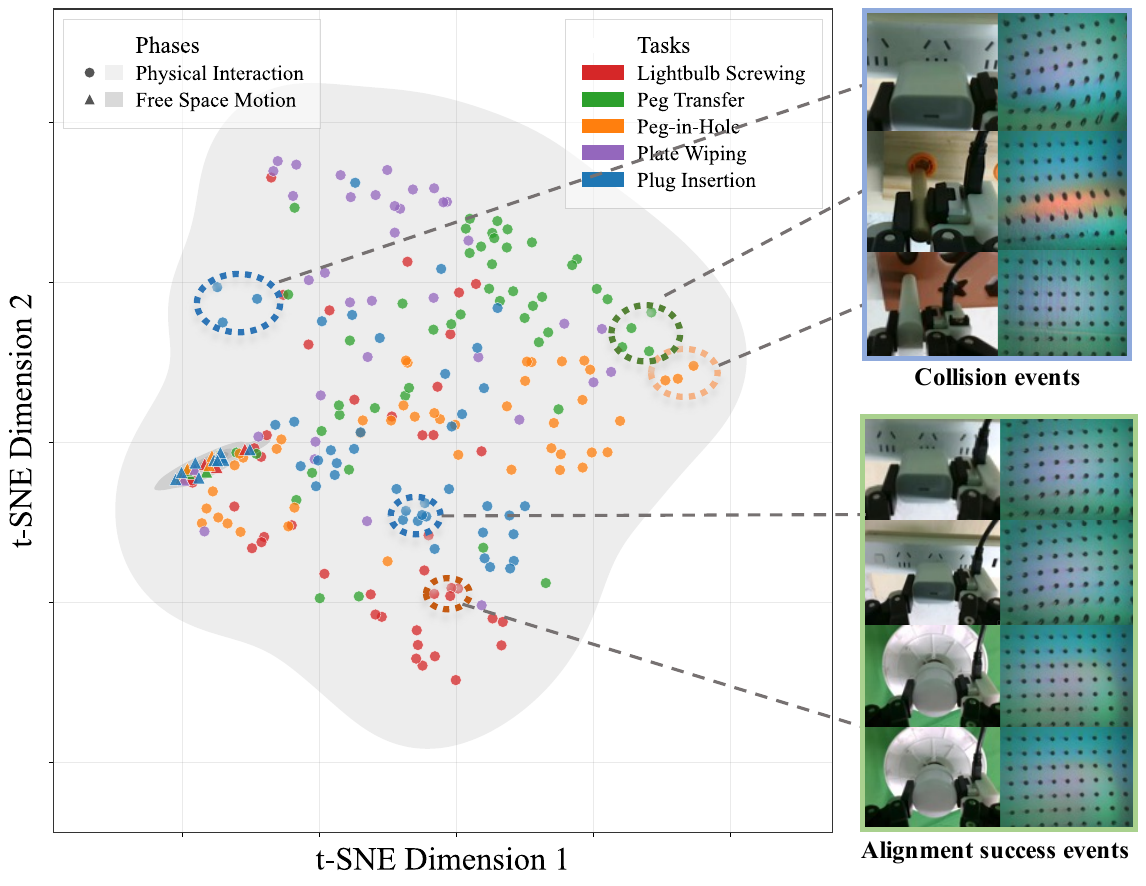}
\caption{\textbf{t-SNE Visualization of Latent Contact Primitives.} Each point represents a per-frame primitive, colored by task identity and annotated by task phase. Free-space motion frames from all five tasks converge into a single, shared compact cluster, reflecting their uniformly low tactile information gain. In contrast, physical interaction frames disaggregate into task-specific local clusters organized around semantically distinct contact events, such as unexpected collision and alignment success.}
  \label{fig:primitives}
\vspace{-6mm}
\end{figure}

\textbf{Emergent Semantics of Latent Contact Primitives.} As shown in Fig.~\ref{fig:primitives}, we investigate the structure of the learned VQ codebook by projecting per-frame latent contact primitives into a 2D space via t-SNE~\cite{vandermaaten2008tsne}, with each frame annotated by both its interaction phase (free-space motion vs.\ physical contact) and task identity. Two salient structural properties emerge after joint pre-training across all five tasks. First, primitives corresponding to free-space motion---the dominant phase of every episode---converge into a single, compact cluster that is shared uniformly across all tasks, reflecting the inherently low tactile information gain during non-contact locomotion where visual priors are highly predictive. Second, during phases of physical interaction, the representation space disaggregates into a set of task-specific local clusters organized around semantically distinct contact events, including unexpected collision and alignment success.

This two-level organization---a global collapse for uninformative phases and a fine-grained task-specific partition for critical contact events---confirms that the Residual Tactile Representation successfully distills continuous, high-dimensional sensory streams into a compact, interpretable vocabulary of contact primitives, providing structured semantic abstractions that are both physically grounded and generalizable across diverse manipulation scenarios.

\begin{figure}[h]
  \vspace{2mm}
  \centering
  \includegraphics[width=.8\columnwidth]{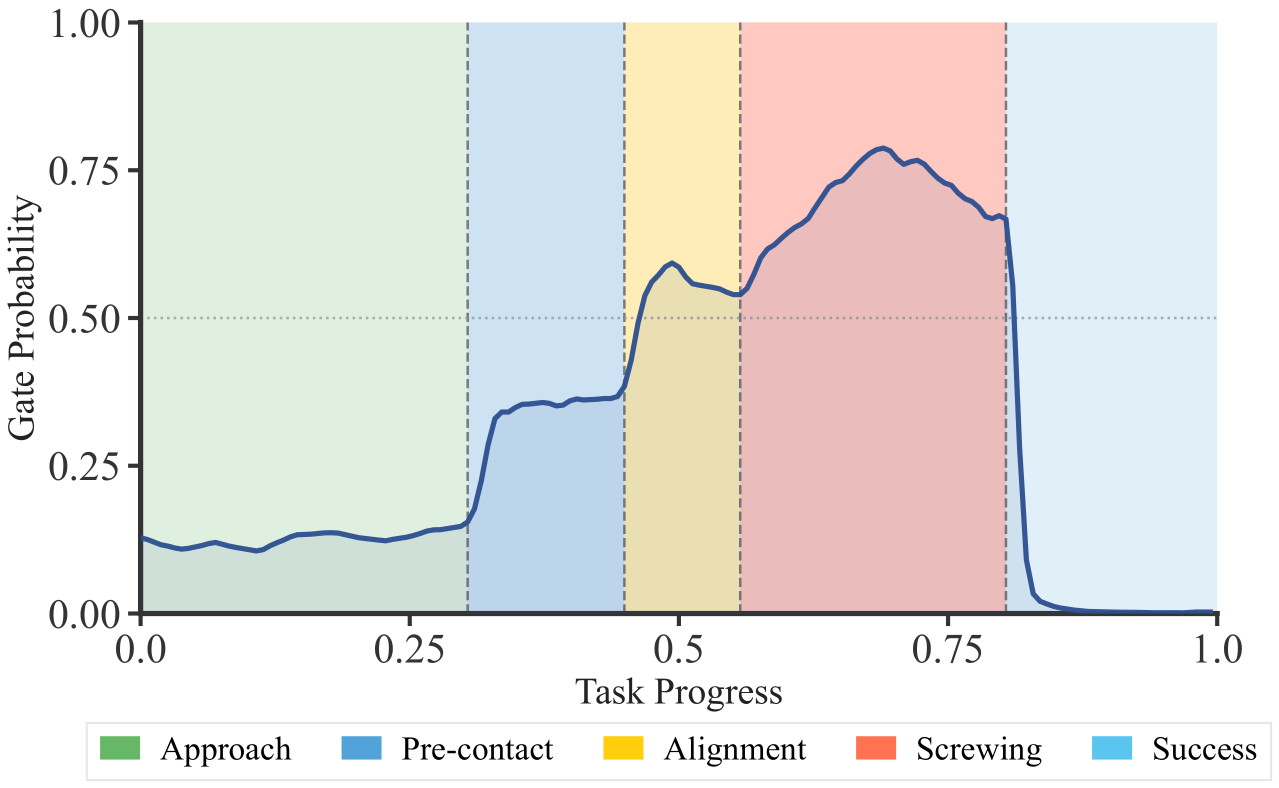}
  \caption{\textbf{Temporal Evolution of the Surprise-Aware Gate $g_t$ During a \textit{Lightbulb Screwing} Trial.} The gate value is shown across five task phases: \textit{Approach}, \textit{Pre-contact}, \textit{Alignment}, \textit{Screwing}, and \textit{Success}. $g_t$ remains near zero during free-space motion and rises progressively as physical contact and thread engagement are established, validating that the gate learns to prioritize tactile feedback precisely when visual predictions become unreliable.}
  \label{fig:gate}
\vspace{-2mm}
\end{figure}

\textbf{Adaptive Regulation via Surprise-Aware Gating.} We further analyze the temporal evolution of the surprise-aware gate $g_t$ during a successful \textit{Lightbulb Screwing} trial (Fig.~\ref{fig:gate}). The curve demonstrates that the gating mechanism naturally learns a phase-dependent modulation strategy consistent with human sensorimotor control. During the \textit{Approach} phase (no contact), $g_t$ remains attenuated (near zero), effectively suppressing tactile noise to allow the high-bandwidth visual policy to dominate trajectory planning. Conversely, upon physical contact establishment, $g_t$ rapidly saturates, amplifying the influence of the tactile pathway. This behavior validates our predictive coding formulation: the policy actively gates tactile integration based on visual uncertainty, prioritizing tactile feedback only when it provides critical information gain.

\vspace{-2mm}
\subsection{Ablation Studies}
\label{sec:ablation}

To validate the architectural design of ResTacVLA, particularly the Residual Quantization of the Cross-Modal Predictor and the Surprise-Aware Gating, we conducted comprehensive ablation studies on two representative tasks: \textit{Plug Insertion} and \textit{Plate Wiping}. The results are shown in Table \ref{tab:ablation}.

\textbf{Latent Primitives vs. Continuous Residuals.} We first investigate the contribution of the Vector Quantization (VQ) module by replacing the discrete Latent Contact Primitives with continuous, raw residual embeddings. As shown in Table \ref{tab:ablation}, this ablation results in a noticeable performance degradation (-26.7\%). Qualitatively, this variant exhibits more frequent sporadic jitters and high sensitivity to grasp pose variations. This suggests that the VQ bottleneck serves as a critical information filter, distilling high-frequency tactile noise into semantic contact events that are more robust and generalizable for policy learning.

\textbf{Surprise-Aware Gating vs. Fixed Fusion.} We further analyze the necessity of the Surprise-Aware Gate. Replacing our adaptive mechanism with a fixed fusion strategy (constant weight of 1, i.e., always-on tactile fusion) leads to a significant drop in success rates (-13.3\%). Specifically, the non-gated variant suffers from trajectory drift caused by tactile noise and initial grasp pose errors. This confirms that adaptive gating is essential for dynamically prioritizing tactile feedback during high-surprise events while preserving visual stability in free space.

\begin{table}[t]
\vspace{1.8mm}
\centering
\renewcommand{\arraystretch}{1.2}
\caption{\textbf{Ablation Analysis of the ResTacVLA Architectural Components.}}
\label{tab:ablation}

\begingroup
\renewcommand\TPTminimum{\linewidth}
\begin{threeparttable}
\centering

\resizebox{0.7\linewidth}{!}{%
\begin{tabular}{lcc}
\toprule
\rowcolor{gray!8}
\textbf{Configuration} & \textbf{Avg. Success (\%)} & \textbf{$\Delta$}$^*$ \\
\hline
\textbf{ResTacVLA (Full)} & \textbf{60.0} & -     \\
w/o VQ (Continuous)$^1$       & 33.3          & -26.7 \\
w/o Gating (Fixed)$^2$        & 46.7          & -13.3  \\
$\pi_{0.5}$ (Vision Only)    & 20.0          & -40.0 \\
\bottomrule
\end{tabular}%
}

\begin{tablenotes}[flushleft]
\item[*] \small We evaluate the performance gain $\Delta$ provided by the Residual Tactile Representation. $^1$VQ-based contact discretization, and $^2$Surprise-Aware Gating. Results represent the average success rate (\%) across two representative tasks (\textit{Plug Insertion} and \textit{Plate Wiping}).
\end{tablenotes}
\end{threeparttable}
\endgroup
\vspace{-6mm}
\end{table}


\subsection{Model Robustness}

\begin{figure}[h!t!]
    \centering
    \includegraphics[width=0.8\columnwidth]{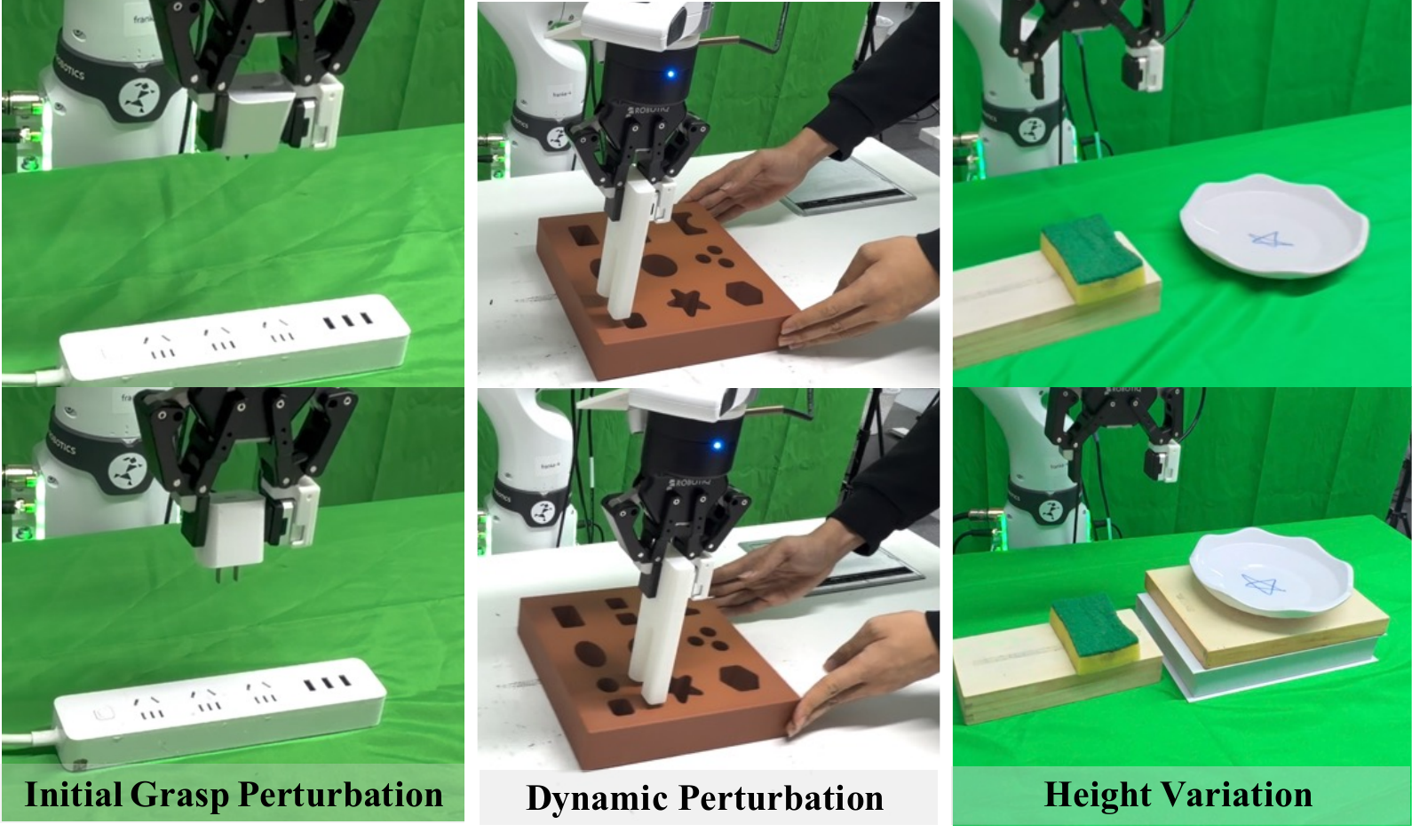}
    \caption{\textbf{Model Robustness Evaluation.} Three perturbation scenarios are designed: (1) initial grasp perturbation applied to \textit{Plug Insertion}, (2) dynamic target displacement during \textit{Peg-in-Hole} execution, and (3) surface height variation applied to \textit{Plate Wiping}.}
    \label{fig:model_robustness}
\vspace{-3mm}
\end{figure}

To further assess the adaptability of ResTacVLA under more challenging and varied conditions, we developed three task-specific robustness evaluations as shown in Fig. \ref{fig:model_robustness}: (1) \textbf{Initial Grasp Perturbation} in \textit{Plug Insertion}, adding translational ($\pm 5$\,mm) and rotational ($\pm 10^\circ$) noise to simulate imperfect grasping conditions; (2) \textbf{Dynamic Perturbation} in \textit{Peg-in-Hole}, applying random target displacements (3--5\,cm) during execution to assess real-time reactivity; and (3) \textbf{Height Variation} in \textit{Plate Wiping}, introducing surface height deviations ($+2$\,cm and $-2$\,cm) to evaluate force compliance.

As shown in Table~\ref{tab:robustness}, across all settings, ResTacVLA exhibited superior generalization, particularly in scenarios requiring fine physical interaction. Under \textit{Initial Grasp Perturbation}, ResTacVLA maintained high success (52.0\%), reflecting its reliance on tactile feedback to compensate for actuation misalignment beyond visual cues. In the \textit{Dynamic Perturbation} setting, it achieved a 66.7\% success rate, outperforming baselines that lacked tactile input or processed it naively. In the \textit{Height Variation} setting, ResTacVLA achieved 53.3\% (+2\,cm) and 40.0\% (-2\,cm) success by effectively scaling its interaction forces to accommodate variable surface depths, avoiding the contact instability and unintended contact loss observed in vision-only models. These results underscore the critical role of the Residual Tactile Representation in intelligently integrating tactile cues---not just for sensing contact, but for modulating action in response to dynamic physical conditions---enabling more versatile and robust robotic manipulation.

\begin{table}[t]
\vspace{3mm}
\centering
\renewcommand{\arraystretch}{1.2}
\caption{\textbf{Robustness Evaluation across Environmental and Actuation Perturbations.}}
\label{tab:robustness}

\begin{threeparttable}

\resizebox{\linewidth}{!}{%
\begin{tabular}{lcccc}
\toprule
\rowcolor{gray!8}
\textbf{Method} & \textbf{Dynamic} & \textbf{Height (+)} & \textbf{Height (-)} & \textbf{Grasp} \\
\hline
$\pi_{0.5}$ (Vision Only)   & 26.7          & 33.3          & 0.0           & 8.0  \\
$\pi_{0.5}$ w/ T-UniT       & 40.0          & 46.7          & 13.3          & 20.0 \\
\textbf{ResTacVLA (Ours)}   & \textbf{66.7} & \textbf{53.3} & \textbf{40.0} & \textbf{52.0} \\
\bottomrule
\end{tabular}%
}

\end{threeparttable}
\begin{minipage}{\linewidth}
\vspace{4pt}
{\tolerance=9999\emergencystretch=3em\small We assess the adaptability of ResTacVLA under dynamic object displacements, significant surface height variations ($\pm 2$\,cm), and initial grasp uncertainties to demonstrate its resilience in unmodeled environments. (Bold: best results)}
\end{minipage}
\vspace{-6mm}
\end{table}

\vspace{-1mm}
\section{CONCLUSIONS}

In this work, we introduced ResTacVLA, a biologically inspired framework designed to efficiently integrate tactile feedback in VLA models for contact-rich manipulation. By reformulating tactile feedback as a residual representation grounded in predictive coding, our approach effectively isolates and amplifies the orthogonal information gain provided by touch. This architecture couples latent contact primitives with surprise-driven gating, ensuring tactile cues are adaptively modulated across distinct task phases. Empirical evaluations across five challenging tasks demonstrate that ResTacVLA achieves state-of-the-art performance, exhibiting superior robustness to unexpected environmental disturbances and actuation uncertainties. These findings underscore the efficacy of residual-based sensor fusion in bridging the gap between high-level semantic planning and fine-grained physical interaction, paving the way for more adaptive and resilient generalist robot policies.









\bibliographystyle{IEEEtran}
\bibliography{references}

\end{document}